\documentclass[fleqn,10pt,twocolumn]{ICCAS2019}

\usepackage{color}

\begin{document}

\title{Deep Reinforcement Learning Based Robot Arm Manipulation with Efficient Training Data through Simulation}

\author{Xiaowei Xing${}^{1}$ and Dong Eui Chang${}^{1*}$ }

\affils{ ${}^{1}$School of Electrical Engineering, Korea Advanced Institute of Science and Technology, \\
Daejeon, 34141, Korea (xwxing@kaist.ac.kr, dechang@kaist.ac.kr) ${}^{*}$ Corresponding author}

\thanks{ \noindent
  This research has been in part supported by Institute for Information \& communications Technology Planning \& Evaluation(IITP) grant funded by the Korea government(MSIT) (No. 2019-0-01396, Development of framework for analyzing, detecting, mitigating of bias in AI model and training data) and by the ICT R\&D program of MSIP/IITP [2016-0-00563, Research on Adaptive Machine Learning Technology Development for Intelligent Autonomous Digital Companion].
  }

\abstract{
    Deep reinforcement learning trains neural networks using experiences sampled from the replay buffer, which is commonly updated at each time step. In this paper, we propose a method to update the replay buffer adaptively and selectively to train a robot arm to accomplish a suction task in simulation. The response time of the agent is thoroughly taken into account. The state transitions that remain stuck at the boundary of constraint are not stored. The policy trained with our method works better than the one with the common replay buffer update method. The result is demonstrated both by simulation and by experiment with a real robot arm. 
}


\maketitle


\section{Introduction}
Traditionally, robot arm control has been handled by inverse kinematics (IK), which makes use of the kinematics equations to determine joint angles that provide a desired position for each of the robot's end-effector.
Since it is not always possible to obtain a closed form solution, iterative optimization is typically relied on to seek out an approximate solution.
Reinforcement learning (RL) has been shown to be an effective framework for a range of robotic control tasks, from locomotion \cite{ref1} to autonomous vehicle control \cite{ref2}.
 However, the high number of degrees-of-freedom of modern robots leads to large dimensional state spaces, from which a desired policy is difficult to be learned.
 Deep reinforcement learning (DRL), which is the conjunction of deep neural network (DNN) \cite{ref3} and RL, approximates non-linear multidimensional functions by parameterizing agents' experiences through finite weights of the network, thereby overcoming the disadvantage caused by a huge or even infinite memory for storing experiences.
DRL makes it possible for robots to be more human-like in some tasks, such as door opening \cite{ref4} and walking \cite{ref5}, even if some interior knowledge of the system may be unavailable to us.

DRL for robot manipulation often uses real-world robotic platforms, which usually require human interaction and long training time, leading to a large amount of resource consumption and potential danger.
A promising alternative solution is to use simulation to produce a controller, which can then be applied to real-world hardware directly or with some minor adjustments, scaling up the available training data without the burden of requiring human interaction during the training process \cite{ref6}.

Building upon the recent success of continuous control based on deep deterministic policy gradient (DDPG) \cite{ref7}, we present an approach that uses DDPG and 3D simulation to train a robot arm in an object sucking task.
For the effectiveness of the training data, we adopt a novel approach to update the replay buffer adaptively, considering response time and joint constraints.
Simulation experiments demonstrate that our approach performs better than the common approach, which is updating the replay buffer at each time step. Real-world experiments also show the good quality of the policy trained with our method.

\section{Background}
\subsection{Preliminaries}
We consider the standard RL framework in which an agent acts in a stochastic environment by sequentially choosing actions over a sequence of time steps in order to maximize a cumulative reward, where we assume that the environment is fully observable.
An environment is modeled as a Markov decision process (MDP) and is defined by a state space $\mathcal{S}$, an action space $\mathcal{A}$, an initial state distribution with probability $p_{1}(s_{1})$, a stationary transition dynamics distribution with probability $p(s_{t+1}|s_{t}, a_{t})$ satisfying the Markov property $p(s_{t+1}|s_{1}, a_{1}, ..., s_{t}, a_{t})=p(s_{t+1}|s_{t}, a_{t})$, for any trajectory $s_{1}, a_{1}, s_{2}, a_{2}, ..., s_{T}, a_{T}$ in a state-action space, and a reward function $r: \mathcal{S}\times\mathcal{A}\rightarrow\mathcal{R}$ \cite{ref7}.
We denote the stochastic policy parameterized by $\theta$ by $\pi_{\theta}: \mathcal{S}\rightarrow\mathcal{{P(A)}}$, where $\mathcal{P(A)}$ is the set of probability measures on $\mathcal{A}$, and  $\pi_{\theta}$ is understood as the map $s_t \mapsto \pi_{\theta}(\cdot |s_{t})$ where $\pi_{\theta}(a_{t}|s_{t})$ is the conditional probability at $a_{t} \in \mathcal{A}$ given $s_t \in \mathcal{S}$. For the deterministic policy, a map $\mu_{\theta}: \mathcal{S}\rightarrow\mathcal{A}$ is used, instead.

The return, defined as the sum of discounted future rewards from time-step $t$ onwards, is denoted as $R_{t}=\sum_{i=t}^{T}\gamma^{i-t}r(s_{i}, a_{i})$ with a discounting factor $\gamma\in[0, 1]$. The goal in RL is to learn a policy which maximizes the expected return from the initial state $J=\textbf{E}_{s_{i>0},a_{i>0}}[R_{1}]$.
The action-value function is defined to be the expected return after taking an action $a_{t}$ at state $s_{t}$ as follows:
\begin{align*}
&Q(s_{t},a_{t})=\textbf{E}_{s_{i>t},a_{i>t}}[R_{t}|s_{t},a_{t}]. 
\end{align*}

\subsection{Policy gradient}
For a stochastic parametric policy $\pi_{\theta}$, the objective is to find the optimal parameter $\theta^{*}$ that maximizes the expected return $\theta^{*}=\operatorname{argmax}_{\theta}J$. The basic idea behind policy gradient algorithms is to adjust the parameter $\theta$ in the direction of the gradient $\nabla_{\theta} J$. The fundamental theorem underlying these algorithms is the policy gradient theorem \cite{ref8} shown by 
\begin{align}
&\nabla_{\theta}J(\pi_{\theta})=\textbf{E}_{s,a}[\nabla_{\theta}\log\pi_{\theta}(a|s)Q^{\pi_\theta}(s,a)]. \label{eq.2}
\end{align}
The policy gradient theorem is of great significance since the computation of $\nabla_{\theta} J$ is replaced by a simple expectation. 
If the target policy is deterministic \cite{ref9}, the policy gradient theorem is in the form of Eq. \eqref{eq.3}.
\begin{align}
&\nabla_{\theta}J(\mu_{\theta})=\textbf{E}_{s}[\nabla_{\theta}\mu_{\theta}(s)\nabla_{a}Q^{\mu_{\theta}}(s,a)|_{a=\mu_{\theta}}]. \label{eq.3}
\end{align}
In our framework, deterministic policy gradient is applied to tackle the continuous high-dimensional action space.

\subsection{Actor-critic algorithms}
The actor-critic is a popular layout based on the policy gradient theorem. An actor adjusts the parameters of $\pi_{\theta}$ or $\mu_{\theta}$ by Eq. \eqref{eq.2} or Eq. \eqref{eq.3}.
A critic approximates the state-action value function by $Q_{\omega}(s,a)$, which is parameterized by $\omega$ and replaces the true $Q^{\pi}(s,a)$ in Eq. \eqref{eq.2} or $Q^{\mu}(s,a)$ in Eq. \eqref{eq.3}.

Typically a parameterized family of policies are applied to actor-only methods with policy gradient algorithms, which suffer from high variance in the estimates of the gradient \cite{ref10}.
Critic-only methods that use temporal difference learning have a lower variance in the estimates of expected rewards \cite{ref11}, where a straightforward way to derive a policy is by selecting actions greedily. However, one needs to select an action of the highest value in every state encountered, which can be computationally intensive.
Actor-critic methods take advantage of their virtues. While the critic supplies the actor with low-variance estimates of the performance, the parameterized actor provides actions directly, without the procedure of finding the best action at each state.

\subsection{Experience replay}
Experience replay, first introduced by Lin \cite{ref12}, enjoys a great success recently in the DRL community and has become a new norm in many DRL algorithms \cite{ref13}. The key idea of experience replay is to train the agent with ``experiences" sampled from the replay memory, which is also called a replay buffer.
An experience is defined as a tuple $(s_{t},a_{t},r_{t},s_{t+1})$.
At each time step, the current transition is added to the replay buffer and some experiences are sampled randomly as a mini-batch. Uniform sampling is the simplest and a popular strategy, while prioritized sampling \cite{ref14} is proposed to assign relative priorities to items in the buffer.
When the replay memory is full, the earliest experience is eliminated to make room for the latest one.
Experience replay not only provides uncorrelated data for training \cite{ref15}, but also significantly improves the data efficiency \cite{ref16}, which is a desired property to alleviate the data-hungry problems for many DRL algorithms.

\section{Method}
Our objective is to train policies for a robot arm with simulation that can perform a task in both simulation and real world. The frequency of updating replay memory is adapted to record steady state responses of the arm to the joint commands.
Those state transitions that remain stuck at the boundary of constraint are not used to update the buffer.

\subsection{Tasks}
\begin{figure}[thb]
\begin{center}
\includegraphics[width=6.5cm]{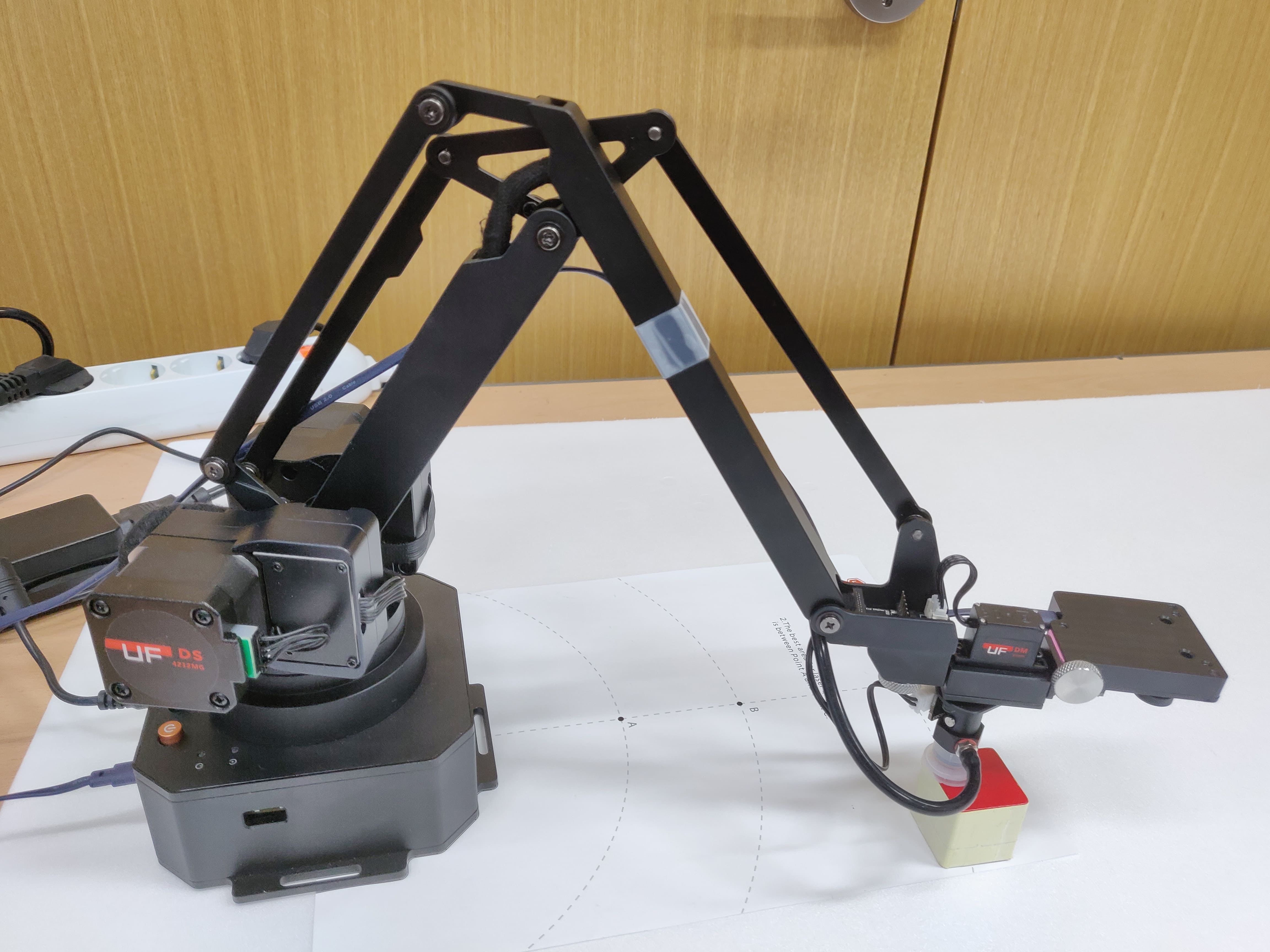}
\caption{\label{iccas1} uArm Swift Pro.}
\label{fig:iccas1}
\end{center}
\end{figure}

\begin{figure}[thb]
\begin{center}
\includegraphics[width=6.5cm]{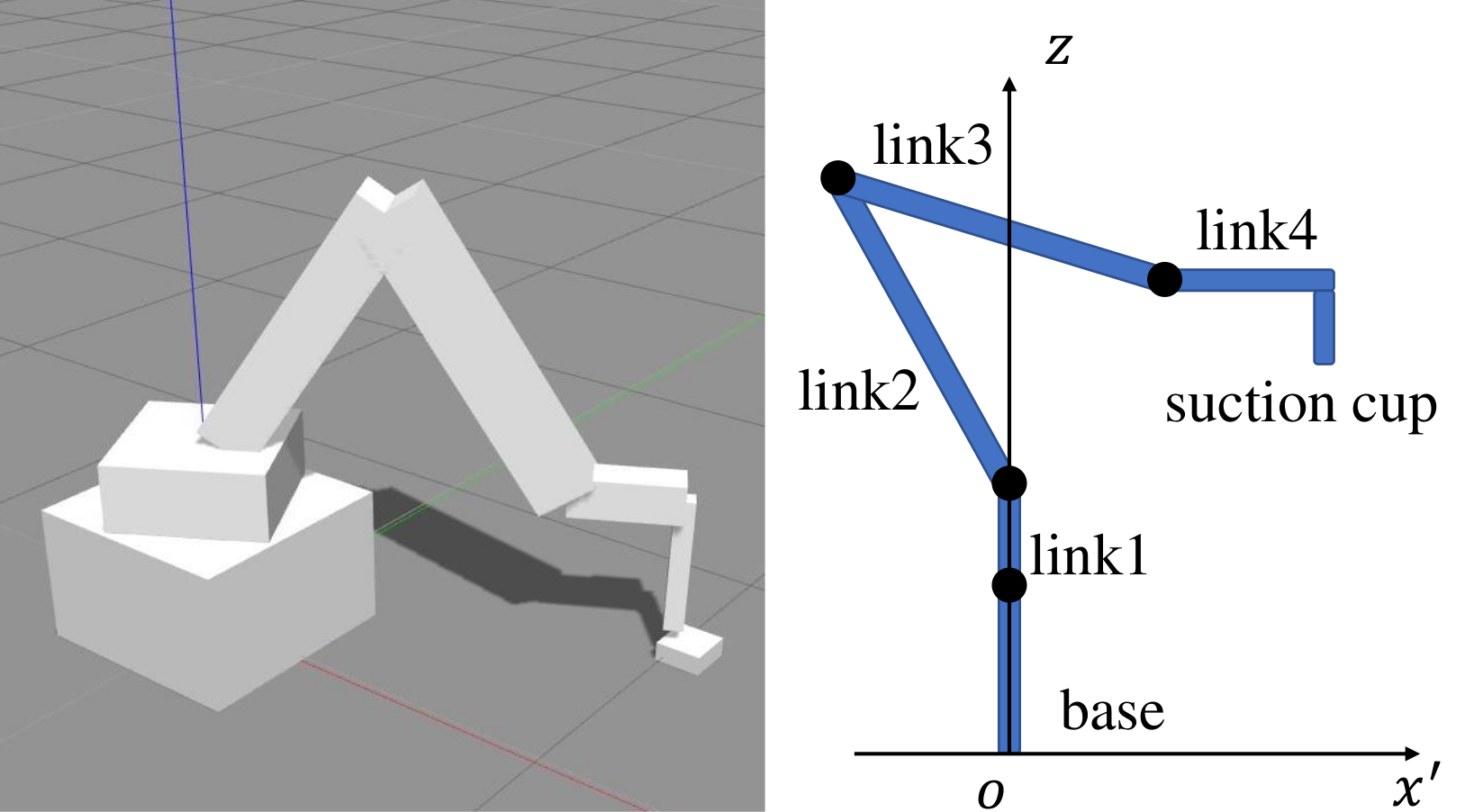}
\caption{\label{iccas2} Simulation model in Gazebo and simplified mathematical model.}
\label{fig:iccas2}
\end{center}
\end{figure}

Our experiments are conducted on an object sucking task using a uArm Swift Pro robot arm. There are three motors installed to control three key joints according to which other joints change their positions. The simulation model is simplified to be the series connection instead of the original closed-loop structure. Images of the real robot and the simulation model  are shown in Fig.~\ref{fig:iccas1} and Fig.~\ref{fig:iccas2}.
At the start of each episode, joint positions are initialized with default angles and the position of object is set randomly in the operating range of the suction cup, which is the end-effector of the arm. The goal in real world is to suck the object, while in simulation it is set to move the suction cup within a certain distance to the central point of the upper surface of the object.

\subsection{State and action}
The state is represented using positions of the end of links and the suction cup, the distance between the suction cup and the object, joint positions, and a flag that is turned on if  the suction cup is close enough to the object for a continuous period of time. The turn on of the flag indicates the end of episode. 
The combined features result in a 20D state space.
Due to the structure of original arm, three joint position commands are needed to specify the position of end-effector. Thus, for the simulation model, three angle signals are enough for arm control. A 3D action space is yielded by the policy, which instructs differences between reference and current joint positions.

\subsection{Replay buffer update}
Any physical system requires a certain period of time to reach a state expected by the command even though this period may be very short. The most widely used method is to update the replay buffer at each time step. Thus, it can be referred as uniform memory update. Compared with this method, we propose to store a new state when the simulation model reaches a steady state, which means that joint position differences between last and current time step are less than a certain threshold.
Those state transitions that remain stuck at the boundary of constraint are not added to the reply buffer because they do not provide any useful experience for training.


\subsection{Reward}
We use the following reward function
\begin{align}
r(s_{t},a_{t})&=\alpha r_{d}(f(s_{t},a_{t}))+\beta (r_{d}(f(s_{t},a_{t}))-r_{d}(s_{t})) \notag \\
&\quad+\sigma d_{o}(s_{t},f(s_{t},a_{t})), \label{eq.4}
\end{align}
where $f(s_{t},a_{t})$ is determined by the system and gives the next state $s_{t+1}$; $r_{d}(s_{t})=\alpha_{d}d_{h}(s_{t})+\beta_{d}d_{v}(s_{t})$ with $\alpha_d<0$ and $\beta_d<0$; $d_{h}(s_t)$ and $d_{v}(s_t)$ are the horizontal and vertical distances from the suction cup to the object, respectively; $d_{o}(s_t,f(s_{t},a_{t}))$ is the position displacement of the object between $s_{t}$ and $s_{t+1}$; and $\alpha>0$, $\beta>0$ and $\sigma<0$ are scalar factors.  
The reward function assesses each action as follows. The effect of the first term in Eq. \eqref{eq.4} is that if the distance between the suction cup and the object is short in $s_{t+1}$, then the action is regarded to be profitable. The second term shows that if the suction cup is closer to the target in $s_{t+1}$ than in $s_{t}$, $a_{t}$ is considered to be beneficial. The last term punishes the action if a change of the target position appears.

\subsection{Network and learning algorithm}
We use an actor-critic learning framework for our experiments, mainly adapted from DDPG. The actor $\mu_{\theta}$ and the critic $Q_{\omega}$ are estimated by fully connected neural networks, parameterized by $\theta$ and $\omega$, respectively. ReLU activation function is used after each hidden layer, and the output of the actor is passed through a tanh activation function to limit the ultimate output, which instructs the differences between target and present joint positions.
To stabilize the training, target networks $\mu_{\theta'}$ and $Q_{\omega'}$ are created, which are copies of the actor and critic networks. The target networks $\mu_{\theta'}$ and $Q_{\omega'}$ take next state as input variable, with their parameters $\theta'$ and $\omega'$ being updated slowly to track the learned networks.
A mini-batch consisting of experience tuples $(s_{t},a_{t},s_{t+1},r_{t})$ is fed to training at each time step.
The critic is optimized by minimizing the loss represented by 
\begin{align*}
&L(\omega) =\frac{1}{N}\sum_{i}(y_{i}-Q_{\omega}(s_{i},a_{i}))^{2},
\end{align*}
where $y_{i}=r_{i}+\gamma Q_{\omega'}(s_{i+1},\mu_{\theta'}(s_{i+1}))$. The actor is updated with the policy gradient in Eq. \eqref{eq.3}. The target networks are updated by $\theta'\leftarrow\tau\theta+(1-\tau)\theta'$ and $\omega'\leftarrow\tau\omega+(1-\tau)\omega'$.

\section{Experiments}

\subsection{Simulated experiments}
We evaluate our method by means of simulated tasks in the Gazebo physics engine, where fast comparisons of design choices (e.g., update frequencies, network architectures, and other hyperparameters) are possible. The whole system, including neural networks and the arm model, is synthesized in  Robot Operating System (ROS).

The robot arm model is simplified into an open-loop model consisting of necessary links and joints. Since the suction cup is always vertical to the ground and is fixed on a link with an angle of $90^\circ$, five links are enough to describe the real arm.
The model in Gazebo, comprised of base, link1, link2, link3 and link4 jointed by joint1, joint2, joint3 and joint4, and the simplified mathematical model are shown in Fig.~\ref{fig:iccas2}, where link4 is horizontal and the coordinate plane $ox'z$ is spanned by link2 and link3. Given the length of each link, the position of the end of a link can be calculated with joint angles which are returned from Gazebo. Together with the position of the object, the state is well defined. The action belonging to the 3D space sends commands to joint1, joint2 and joint3 from which the command for joint4 is  automatically determined.

The actor network and the critic network both have two hidden layers containing 400 and 300 perceptions, and so do the target actor network and the target critic network.
The discounting factor is set to $\gamma=0.9$  and the ADAM optimizer \cite{ref17} is used to update parameters, with stepsize $1\times10^{-4}$ for the actor and $1\times10^{-3}$ for the critic.
The size of mini-batch is set to 128.
To explore more possibilities in action space, an OU noise is applied to the output of the actor network.
To balance between exploration and exploitation, the amplitude of noise is gradually reduced as the training proceeds.

\begin{figure}[thb]
\begin{center}
\includegraphics[width=7.5cm]{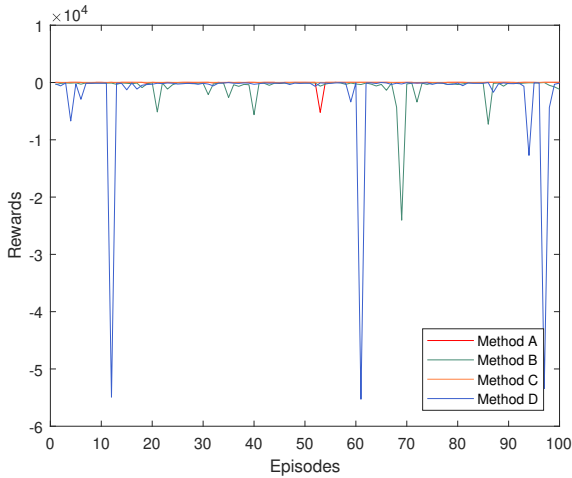}
\caption{\label{iccas4}Rewards of different methods during testing.}
\label{fig:iccas3}
\end{center}
\end{figure}

To verify the effectiveness of our data-collecting approach and reward function, we train policies with four different methods as follows:
\begin{itemize}
\item[A.] Update the replay buffer adaptively (ours) + apply no position change penalty ($\sigma =0$ in Eq. \eqref{eq.4}),
\item[B.] Update the replay buffer at each time step (uniform memory update) + apply no position change penalty ($\sigma =0$ in Eq. \eqref{eq.4}),
\item[C.] Update the replay buffer adaptively (ours) + apply position change penalty ($\sigma <0$ in Eq. \eqref{eq.4}),
\item[D.] Update the replay buffer at each time step (uniform memory update) + apply position change penalty ($\sigma <0$ in Eq. \eqref{eq.4}).
\end{itemize}
During training, the simulation frequency is set to 20Hz and 1500 episodes are carried out. In each episode 200 experiences are stored in the replay buffer. For a fair comparison, we generate 100 random initial positions of the object and use these initial conditions to test each learned policy. Rewards and success rates of these tests are illustrated in Fig.~\ref{fig:iccas3} and Table~\ref{table:table1}.

\begin{table}[htb]
\setlength{\extrarowheight}{0.75ex}
\caption{Success rate of each method.}
\label{table:table1}
\begin{center}
\begin{tabu}to\linewidth{|X[c]|X[0.5c]|X[0.5c]|X[0.5c]|X[0.5c]|}\hline
Method &   A   &   B   &   C   &   D \\\hline
Success rate & 99\% & 75\% & 100\% & 80\% \\\hline
\end{tabu}
\end{center}
\end{table}

From testing we find that policies trained through the uniform memory update method (B and D) make the arm keep shaking. The reason is that these policies only focus on the robot state in next time step instead of the steady state given a joint command. With our adaptive replay memory update method, the robot performs much more stable. From Fig.~\ref{fig:iccas3} and Table~\ref{table:table1} we can see that both method A and C behave better than method B and D. The success rate of the policy trained through method C is slightly higher than the one trained through method A, testifying that our reward function helps improve the policy.
Results can be seen in the supplemental video located at https://www.youtube.com/watch?v=hjqetMpDr48.

\subsection{Real-world experiments}
For real-world experiments, one of the biggest differences compared with simulation experiments is that the information of the object cannot be obtained directly. An OpenMV camera is installed near the end-effector, whose captured images are processed on a PC which connects the arm. Since object detection is not the main topic of this paper, a quite recognizable object is used for testing. The target position concerning to the robot arm is calculated with a series of coordinate transformations, where pixel coordinate system, image coordinate system, camera coordinate system and robot arm coordinate system are applied.

Due to the quite unstable behaviours in simulation with the uniform memory update method, only the policy generated by our data-collecting method and reward function is tested. 50 tests are run with a success rate of 80\%. The result can be seen in the supplemental video located at https://www.youtube.com/watch?v=lUaKoURbZkM.

\section{Conclusions and future work}
We demonstrate the effectiveness of our approach to updating the replay buffer. The replay buffer is updated until the agent completely responds to the input signal.
Those state transitions that remain stuck at the boundary of constraint are not added to the reply buffer. Simulation and real-world experiments show good quality of our method.

Obstacles are not considered in this paper. To train a policy that helps the robot avoid interference while accomplishing tasks automatically will undoubtedly release more labour and enhance productivity. The commands produced by the policy are on the level of angle.
It can be beneficial to control the robot arm on the level of torque.
Exploring these topics with our method could enable more efficient training processes and higher-quality neural network policies in future work.

%

%

\end{document}